\documentclass[letterpaper, 10 pt, conference]{ieeeconf}  % Comment this line out if you need a4paper

\IEEEoverridecommandlockouts                              % This command is only needed if 
\usepackage[pdftex]{graphicx}
\graphicspath{{../pdf/}{../jpeg/}}
\DeclareGraphicsExtensions{.pdf,.jpeg,.png}
\usepackage[cmex10]{amsmath}
\usepackage{amssymb}
\usepackage{algorithmic}
\usepackage{array}
\usepackage{mdwmath}
\usepackage{mdwtab}
\usepackage{eqparbox}
\usepackage{url}
\usepackage[utf8]{inputenc}
\usepackage{siunitx}
\usepackage{multirow}
\title{\bf Realistic Ultrasonic Environment Simulation Using Conditional Generative Adversarial Networks}

 \author{Maximilian Pöpperl, Raghavendra Gulagundi, Senthil Yogamani and Stefan Milz\\Valeo Schalter und Sensoren GmbH\\Kronach, Germany\\maximilian.poepperl@valeo.com}

\begin{document}

\maketitle
\thispagestyle{empty}
\pagestyle{empty}

%%%%%%%%%%%%%%%%%%%%%%%%%%%%%%%%%%%%%%%%%%%%%%%%%%%%%%%%%%%%%%%%%%%%%%%%%%%%%%%%

\begin{abstract}
Recently, realistic data augmentation using neural networks especially generative neural networks (GAN) has achieved outstanding results. The communities main research focus is visual image processing. However, automotive cars and robots are equipped with a large suite of sensors  to achieve a high redundancy. In addition to others, ultrasonic sensors are often used due to their low-costs and reliable near field distance measuring capabilities. Hence, Pattern recognition needs to be applied to ultrasonic signals as well. Machine Learning requires extensive data sets and those measurements are time-consuming, expensive and not flexible to hardware and environmental changes. On the other hand, there exists no method to simulate those signals deterministically. We present a novel approach for synthetic ultrasonic signal simulation using conditional GANs (cGANs). For the best of our knowledge, we present the first realistic data augmentation for automotive ultrasonics. The performance of cGANs allows us to bring the realistic environment simulation to a new level. By using setup and environmental parameters as condition, the proposed approach is flexible to external influences. Due to the low complexity and time effort for data generation, we outperform other simulation algorithms, such as finite element method. We verify the outstanding accuracy and realism of our method by applying a detailed statistical analysis and comparing the generated data to an extensive amount of measured signals.
\end{abstract}

%%%%%%%%%%%%%%%%%%%%%%%%%%%%%%%%%%%%%%%%%%%%%%%%%%%%%%%%%%%%%%%%%%%%%%%%%%%%%%%%

\section{Introduction}
Ultrasonic sensors are used in a wide field of applications. Although high performance sensor solutions are available in medical applications \cite{medical_ultrasonic}, automotive and robotic scenarios prefer low-cost versions \cite{low_cost_uls}, \cite{robot_nav}. As the production costs are low, ultrasonic sensors are widely spread in such scenarios. However, the low-cost realization restricts the functionality, so that ultrasonic sensors are mainly known as reliable near-range distance sensors \cite{near_range_uls}.

To keep up with the current requirements for autonomous driving and autonomous robot navigation, a simple distance measurement is not sufficient anymore \cite{autonomous_driving}. Further information about the environment, e.g. ground type, and about the reflecting object, e.g. height and width, is required. Simultaneously the low cost advantage of ultrasonic sensors needs to be preserved, so that hardware changes are prevented. Consequently, the only way to improve the performance of those sensors is the development of novel algorithms, e.g. based on machine learning \cite{ml_uls}.

Pattern recognition algorithms basically require a set of labeled data \cite{ml_dataset}. Creating a large data set based on real measurements in different scenarios is time consuming and costly. Furthermore it provides a low flexibility, if minor changes are made in the hardware setup. Hence, synthetically generated data is preferable, as it can be adapted easily and a large amount of ultrasonic signals can be generated with low temporal effort.

\begin{figure}[b]
\centering{\includegraphics[width=88mm]{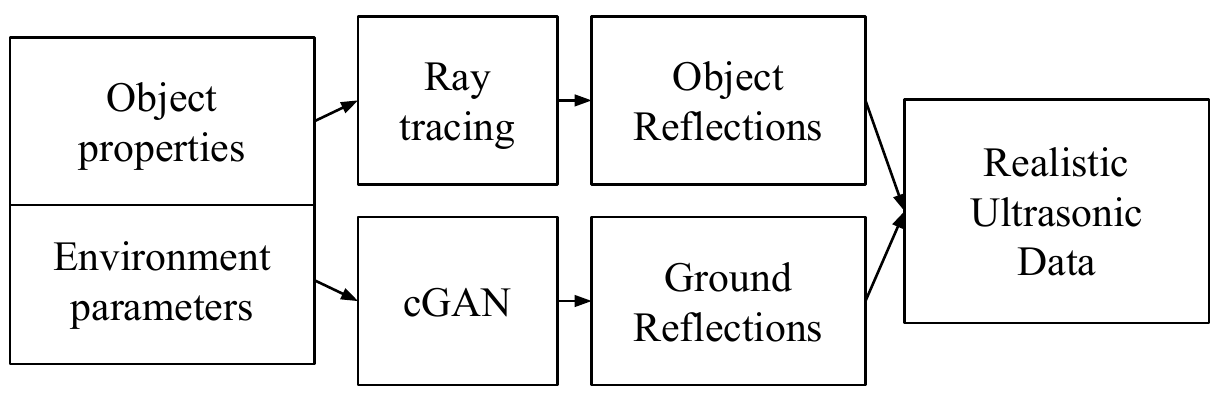}}
\caption{Simulation concept using cGANs to generate ground reflections}
\label{sim_idea}
\end{figure}
The reflections at objects in the environment are satisfactorily modeled by quasi-optical algorithms (e.g. ray-tracing) \cite{ray_tracing}. The modeling of simulations at the ground surface is much more complex. Standard methods, such as finite element method, require a huge amount of storage, processing power and computation time. This is not sensible in order to create large data sets. Statistical methods can overcome this problem. Hence, we propose a two step solution for a complete ultrasonic signal simulation. As exemplarily shown in Fig. \ref{sim_idea} the reflections of objects are generated deterministically, whereas the ground reflections are generated by a statistical algorithm. Using superposition principle, both steps can be merged. 

A possibility of a statistical algorithm for the given purpose is a cGAN \cite{cgan}. cGANs allow a random generation of the ground reflections with respect to particular conditions. Although the training of the neural network is complex, the sample generation is of low complexity \cite{gan}. Thus, the approach is suited well to generate large data sets or simulate ultrasonic sensor data. By using suitable conditions, a flexibility regarding hardware or environmental changes can be achieved.

Consequently, cGANs seem to be a satisfactory approach for ultrasonic ground reflection generation. We evaluate the performance of the network by comparing the statistical properties of the generated data to ones of real measurements. By changing the hardware setup and the environmental conditions, we additionally verify the flexibility of the proposed approach.

%%%%%%%%%%%%%%%%%%%%%%%%%%%%%%%%%%%%%%%%%%%%%%%%%%%%%%%%%%%%%%%%%%%%%%%%%%%%%%%%

\section{Measurements}
In a first step, measurements are conducted. One the one hand measurements are required to train the neural network. For this purpose, we need to cover all relevant scenarios with a sufficient amount of ultrasonic signals. On the other hand the measurements are needed to evaluate the generated signals.

\subsection{Measurement equipment}
\begin{figure}[tb]
\centering{\includegraphics[width=44mm]{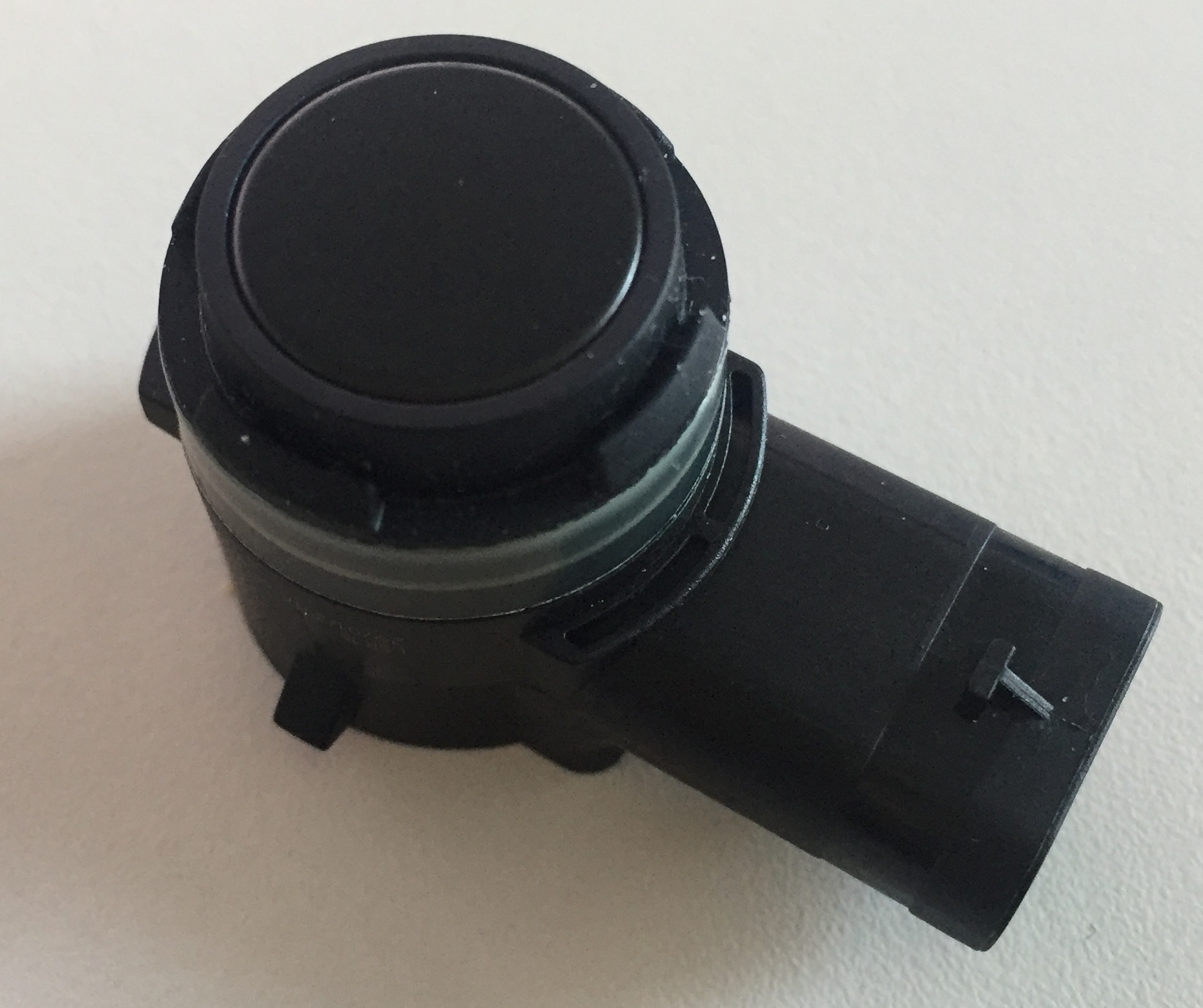}}
\caption{Valeo ultrasonic sensor that is used for measurements}
\label{sensor}
\end{figure}

For the measurements we use a standard ultrasonic sensor that is used for parking assistance. Automotive ultrasonic sensors usually only provide a binary signal output due to limitations in data communications. These binary signals depend on previously set thresholds. Furthermore the 1-bit quantization is lossy and drastically limits the performance of signal processing algorithms. To avoid the dependency on the thresholds and the disadvantages of the binary output, an extended version of the sensor is used that allows the acquisition of the piezoelectric signal. The sensor uses a modulated pulse at a center frequency of \SI{51.2}{\kilo\hertz}. The bandwidth of the pulse is about \SI{3}{\kilo\hertz}. We use a monostatic measurement setup, where the transmitting sensor is also used for receiving the reflected signals. The sensor is shown in Fig. \ref{sensor}.

As a huge amount of measurement data is needed to train the network and to evaluate the data, we use a robot arm to automate the measurements. The DOBOT magician that is used has four rotation axis. This enables different sensor heights above the ground. Furthermore, different angles relative to the ground plane are usable. The angle relative to the ground surface is named as beta angle in the following. In addition, the rotation of the complete robot arm allows to change the filed of view automatically.

\subsection{Measurement setup}
\begin{figure}[tb]
\centering{\includegraphics[width=88mm]{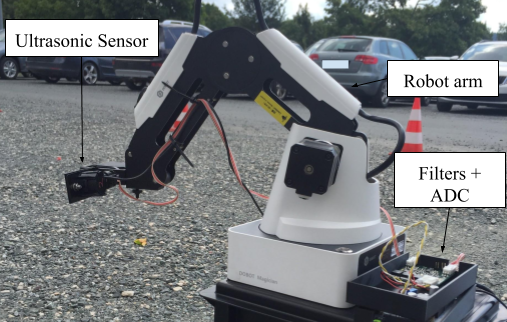}}
\caption{Measurement setup with ultrasonic sensor and robot arm on gravel}
\label{meas_setup}
\end{figure}

To create sensor setups that are similar to automotive scenarios, the sensor is fixed to the robot arm. The robot arm is placed on a platform, as the amplitude of the sensor height is \SI{27}{\centi\meter} only. By using a suitable platform, sensor heights between \SI{35}{\centi\meter} and \SI{62}{\centi\meter} are achieved. The step size in the measurements is \SI{1}{\centi\meter}. The beta angle varies between \SI{-8}{\degree} and \SI{2}{\degree} with a step size of \SI{1}{\degree}. The measurements are done on gravel as well as on asphalt. To achieve a sufficient diversity, we change the field of view of the sensor. By rotating the robot arm by \SI{120}{\degree} using 21 steps, the reflecting ground areas are sufficiently changed. In addition, we conduct 10 measurements at each sensor position and installation angle. These measurement vary due to environmental influences, such as wind and environmental noise. The measurement setup is shown in Fig. \ref{meas_setup}.

In summary, 129,360 measured ultrasonic signals are available for training the neural network and for data evaluation. As ultrasonic data is basically one dimensional and the number of necessary samples per signal is low compared to image processing applications, the amount of measured signals is sufficient to train the networks.

\subsection{Measurement results}
\begin{figure}[b]
\centering{\includegraphics[width=88mm]{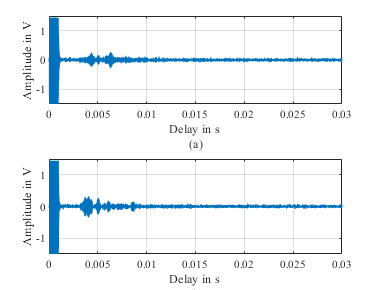}}
\caption{Raw data samples from asphalt (a) and gravel measurements (b) with similar sensor height and installation angle}
\label{raw_data}
\end{figure}
Two samples of the measurements are shown in Fig. \ref{raw_data}. Each signal consists of 9,900 samples with a sampling frequency of \SI{330}{\kilo\hertz}. Oversampling is applied to handle Doppler shift in dynamic scenarios. At the beginning, the reverberation time of the membrane is clearly visible. This is typical for a monostatic ultrasonic setup. Subsequently, the ground reflections appear. The ground reflections vary by changing the measurement setup and the ground type. By comparing Fig. \ref{raw_data} (a) and (b), the differences in the ground reflections become clear. Fig. \ref{raw_data} (a) shows a typical signal on asphalt, whereas Fig. \ref{raw_data} (b) is typical for gravel. For comparability the sensor height and angle are the same in both cases, but similar changes can be observed if the sensor height and angle are changed. To ease the analysis of these effects, it is sensible to process the ultrasonic signal in order to remove the carrier frequency and to improve signal quality.

%%%%%%%%%%%%%%%%%%%%%%%%%%%%%%%%%%%%%%%%%%%%%%%%%%%%%%%%%%%%%%%%%%%%%%%%%%%%%%%%

\section{Data Processing}
The applied processing of the raw measurement data can be divided in two steps. At first a conventional signal processing is applied. This step is also used on the measurement data before training the network, as the signal processing reduces the needed number of samples per measurements as well as the computational effort for the training and the final prediction, respectively. In a second step, the measurement data is analyzed by calculating statistical properties. The statistical properties allow a detailed description of the influence of the hardware setup on the measured signal. Furthermore the statistic analysis can be used to meaningfully compare the synthetic with the measured data.

\subsection{Signal processing}
The aims of the signal processing of the measured signals are on the one hand the improvement of the signal quality and on the other hand the minimization of samples. In order to reduce noise in the signal and remove a possible DC offset, we apply a high order finite response bandpass $s_{b}$:
\begin{equation}
s_{p1} = s_{r}*s_{b}
\end{equation}
The noise reduced signal $s_{p1}$ is mixed to complex baseband afterwards, so that the carrier frequency $f_{c}$ is removed. This process is basic to reduce the sampling frequency drastically and still meet the Nyquist theorem \cite{digital_comm}. The complex baseband signal $s_{p2}$ can be calculated from:
\begin{equation}
s_{p2} = s_{p1}\cdot\exp[-2j\pi f_c t]
\end{equation}
In order to suppress image frequencies and further improve signal quality, a finite impulse response lowpass $s_l$ is applied. The result of the processing steps $s_p$ can be written as
\begin{equation}
s_{p} = s_{p2}*s_{l}
\end{equation}
The processed signal is the complex envelope of the measured raw signal. As the carrier frequency is removed, the required sampling frequency depends on the signal bandwidth and the allowed Doppler shift only. We use a sampling frequency of \SI{20}{\kilo\hertz}. The resulting number of samples per measurement is only 583. This low number of samples is advantageous for training the neural network. If the acceptable Doppler shift is reduced or completely neglected, the number of samples can be further reduced.

The envelope of the signal is still complex valued. To further reduce the complexity of the neural network, we only use the absolute value of the envelope. This is sensible  for most applications, as the analysis of the absolute value is sufficient and the additional information of the complex values are not required. In addition, if the number of quantization bits per sample is limited, the quantization of the absolute value is more effective since there are only positive valued numbers.
\begin{figure}[tb]
\centering{\includegraphics[width=88mm]{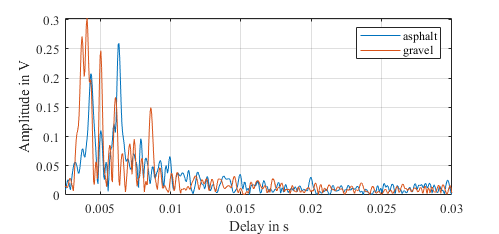}}
\caption{Absolute value of the envelope of the sample signals of Fig. \ref{raw_data} after signal processing}
\label{processed_data}
\end{figure}

\begin{figure}[b]
\centering{\includegraphics[width=88mm]{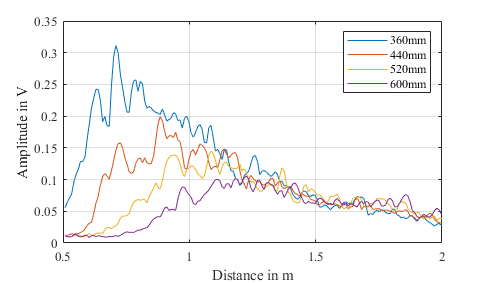}}
\caption{Behavior of the ground reflections for different sensor heights: \SI{36}{\centi\meter} (blue), \SI{44}{\centi\meter} (red), \SI{52}{\centi\meter} (yellow), \SI{60}{\centi\meter} (purple)}
\label{pro_data_height}
\end{figure}

The processing results of the sample measurement signals of Fig. \ref{raw_data} are shown in Fig. \ref{processed_data}. The reverberation time is removed. The differences between the ground reflections can now be seen more clearly. The gravel reflections have more high peaks. The signal range that is mainly affected by the ground reflections is similar. The signal range changes if the sensor height or the beta angle are varied. The behavior of the ground reflections with respect to the sensor height is given in Fig. \ref{pro_data_height}. The plot shows the maximum amplitudes for all measurements with the particular parameters. The amplitude values of the ground reflections decrease with increasing height. In addition, the range, where the maximum of the ground reflections appears, moves to higher distances. This is just because the distance to ground plane increases with increasing sensor height. To describe these observations in detail, a statistical analysis of the data is necessary.

\subsection{Statistical data decomposition}
\begin{figure}[tb]
\centering{\includegraphics[width=88mm]{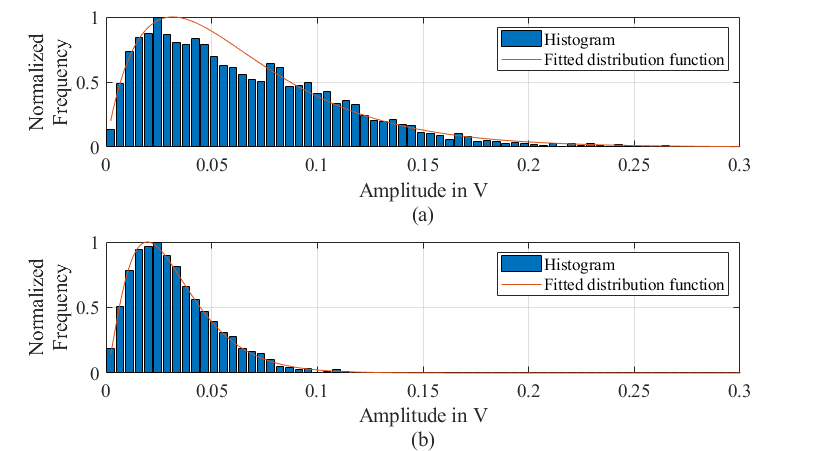}}
\caption{Normalized histograms (blue) and fitted distribution functions (red) for measurements on gravel with a sensor height of \SI{36}{\centi\meter} and a beta angle of \SI{0}{\degree} for a range from \SI{0.5}{\meter} to \SI{0.75}{\meter} in (a) and from \SI{1.5}{\meter} to \SI{1.75}{\meter} in (b)}
\label{histo}
\end{figure}

Since we use statistical methods to describe the reflections of the ground surface, the distribution function of the signal is a starting point for a statistical analysis. As visible in Fig. \ref{processed_data}, the amplitude values differ with varying delay. Thus, evaluating the amplitude distribution of the complete signal is not sensible. We propose to divide the signal into different bins. We choose the bins according to the object distance. Each bin covers a range of \SI{0.25}{\meter}. For each distance bin the amplitude distribution is calculated. Two samples for different distance bins are shown in Fig. \ref{histo}.

The histograms show a characteristic behavior that depends on the measurement setup. This is already visible from Fig. \ref{pro_data_height}. The ground reflections have a distance range, where they mainly appear. In this area the amplitude spread is much higher than in other distance bins of the signal. The sensor height and the beta angle influence the amplitude spread as well as the position of the maximum amplitude frequency. The ground type usually only influences the amplitude spread. To mathematically describe these effects and enable a comparability with synthetic data, we fit a distribution function to the histograms of the measurement data.

Basically the amplitude distribution can be modeled by a Gamma distribution. This has been validated by a $\chi^2$-test. The Gamma distributions allows the description of the amplitude distribution using only two parameters. We use the shape parameter $k$ and the scale parameter $\theta$ to parameterize the Gamma distribution.  

\begin{figure}[tb]
\centering{\includegraphics[width=88mm]{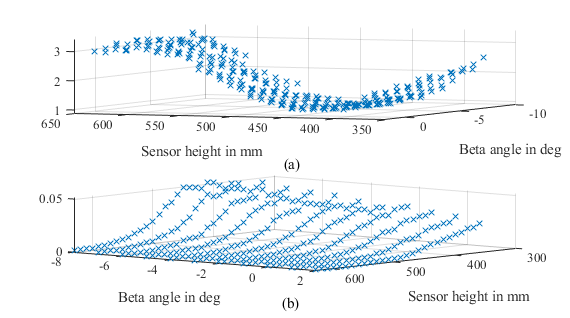}}
\caption{Behavior of the shape parameter $k$ in (a) and of the scale parameter $\theta$ in (b) of gamma distribution for measurements on gravel and a distance bin at \SI{0.5}{\meter} to \SI{0.75}{\meter}}
\label{gamma_param}
\end{figure}

If we evaluate the amplitude distributions of the measurements, a predictable behavior of both parameters is found. The behavior of both parameters for gravel and the distance range of \SI{0.5}{\meter} to \SI{0.75}{\meter} ares shown in Fig. \ref{gamma_param}. A similar behavior can be found for other relevant distance bins and also for asphalt. The predictability of both parameters and the clear tendency allow an interpolation for arbitrary sensor heights and beta angles. Furthermore, this allows a detailed evaluation of synthetically generated data and especially the generalization of machine learning based approaches.

%%%%%%%%%%%%%%%%%%%%%%%%%%%%%%%%%%%%%%%%%%%%%%%%%%%%%%%%%%%%%%%%%%%%%%%%%%%%%%%%

\section{Generative Adversarial Network}
\begin{figure}[b]
\centering{\includegraphics[width=88mm]{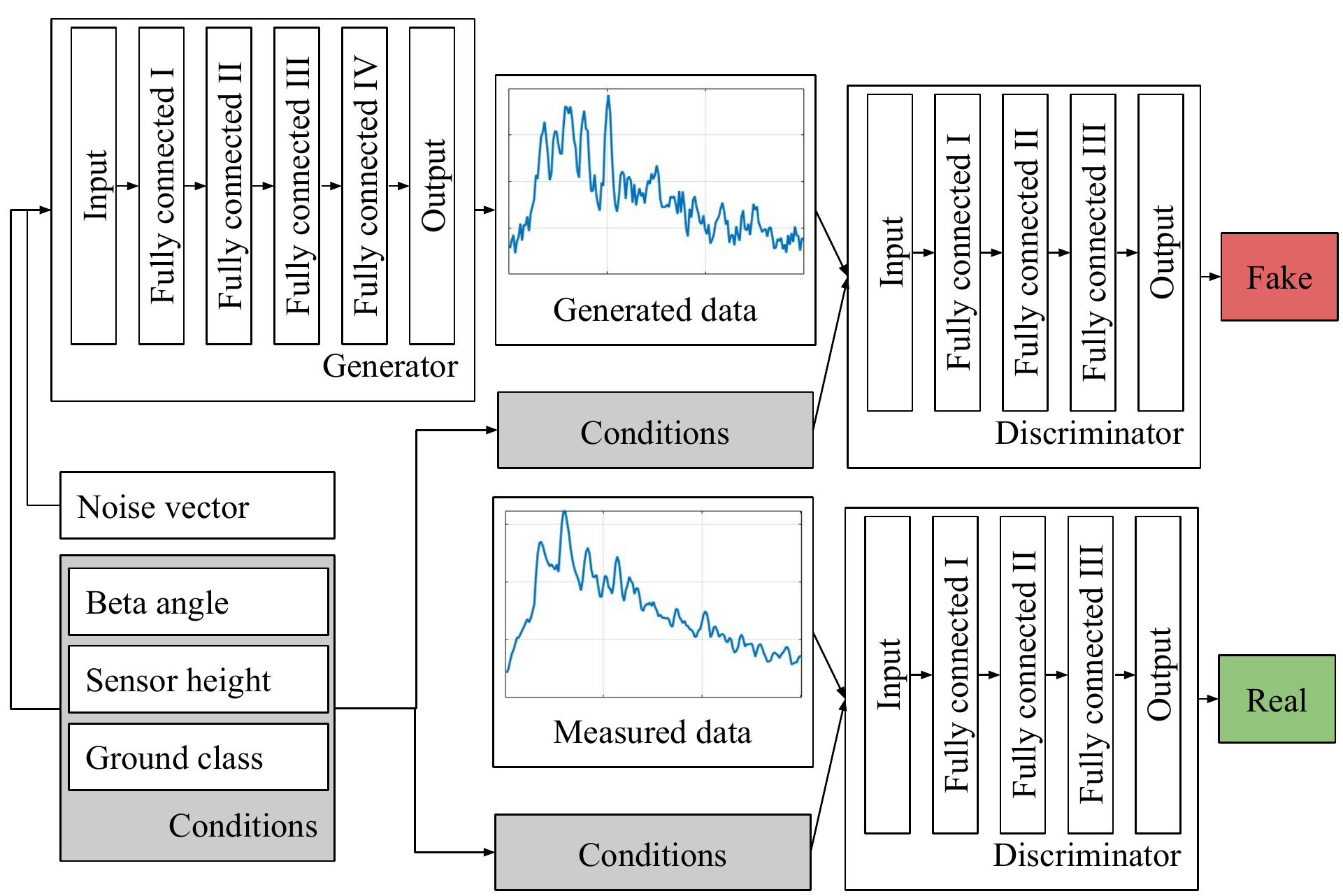}}
\caption{Concept of training the cGAN and draft of the architecture of the generator and discriminator network}
\label{architecture}
\end{figure}
GANs are one possibility to train generative neural networks. In contrast to other generative methods, e.g. deep belief neural networks \cite{deep_belief} or deep Boltzmann machines \cite{deep_boltzmann}, GANs provide different advantages while training as well as in generation performance. Furthermore GANs allow the usage of an additional condition. This means, beside a straightforward data generation, GANs are able to consider additional inputs, so that the generated output can be adapted to certain scenarios. We use the sensor height, beta angle and ground type as condition. Thus we are able to handle all variations of the measurements.
\subsection{Architecture}
An overview of the used network architecture is shown in Fig. \ref{architecture}. To train the cGAN, two separate neural networks are used. Of course a generative neural network $G$ is necessary. We use a simple multilayer perceptron (MLP) for this purpose. The network consists of 4 fully connected layers that use a leaky rectified linear unit (ReLU) as activation function and a fully connected output layer with a hyperbolic tangent as activation function. The network takes a random noise vector $z$ and a conditional vector $x$ as input. The conditional vector is a 1$\times$3 vector that is concatenated to the random vector. The height and angle are both normalized, whereas the ground type is provided as boolean value. We use a normalized signal vector as output $y$. The normalization is needed due to the values range of the hyperbolic tangent. 

The discriminative network $D$ that is needed for training is also a MLP that consists of 3 fully connected layers using a leaky ReLU activation function and a fully connected output layer with a softmax activation. 

During training the generator tries to generate realistic ultrasonic signals based on the random noise vector and the given condition, as the discriminator is trained to distinguish the generated from real measurements. While the generator tries to fool the discriminator with more and more realistic signals, the capability of the discriminator to detect generated signals also increases. Thus, the aim of the training is to minimize the loss of the generator and to maximize the loss of the discriminator. Due to the well-known problem, we can use a standard objective for cGANs \cite{pix2pix}:
\begin{equation}
L = \mathbb{E}_{x,y}\left[\log D\left(x, y\right)\right] + \mathbb{E}_{x,z}\left[\log D\left(x, G\left(x, z\right)\right)\right]
\end{equation}
As described above, the training of the generator results in:
\begin{equation}
G^* = \arg\min_G\max_D L
\end{equation}
The ultrasonic signal vectors are of size 1$\times$583 only. Thus, 100 iterations are sufficient to ensure that the network converges. The implementation of the cGAN in done using PyTorch \cite{pytorch}. 

%%%%%%%%%%%%%%%%%%%%%%%%%%%%%%%%%%%%%%%%%%%%%%%%%%%%%%%%%%%%%%%%%%%%%%%%%%%%%%%%

\section{Validation of Simulation Data}
\begin{figure}[tb]
\centering{\includegraphics[width=88mm]{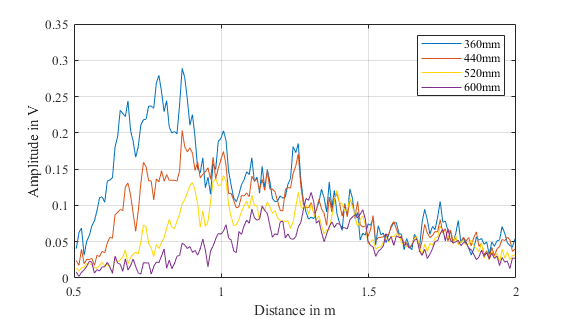}}
\caption{ Behavior of the generated ground reflections for different sensor heights: \SI{36}{\centi\meter} (blue), \SI{44}{\centi\meter} (red), \SI{52}{\centi\meter} (yellow), \SI{60}{\centi\meter} (purple)}
\label{synthetic_data}
\end{figure}

After training the neural network, we can just use the generator network to quickly synthesize ultrasonic signals with ground reflections. Few examples for a gravel scenario and a beta angle of \SI{0}{\degree} are shown in Fig. \ref{synthetic_data}. The generated signals show a similar course as the signals in Fig. \ref{pro_data_height}. But the differences between the signals in both figures also verify that the network generates novel ultrasonic signals. Furthermore the behavior of the ground reflections is modeled correctly using the cGAN, as the amplitude decreases and the range is shifted with increasing sensor height. A difference between the generated data and the measurement data, is the available signal bandwidth. The generated signals show a higher signal bandwidth in contrast to the measurement data. However, this problem can be solved by applying a lowpass filter. To further evaluate the generated data, we apply the statistical processing.

\begin{figure}[b]
\centering{\includegraphics[width=88mm]{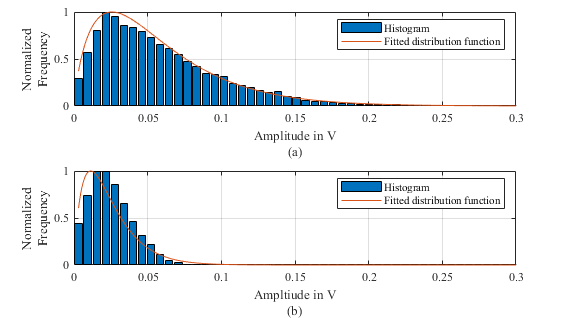}}
\caption{Normalized histograms (blue) and fitted distribution functions (red) for generated data on gravel with a sensor height of \SI{36}{\centi\meter} and a beta angle of \SI{0}{\degree} for a range from \SI{0.5}{\meter} to \SI{0.75}{\meter} in (a) and from \SI{1.5}{\meter} to \SI{1.75}{\meter} in (b)}
\label{synthetic_histo}
\end{figure}

Fig. \ref{synthetic_histo} shows the histogram for the same parameters as in Fig. \ref{histo}. The envelope of the histogram still allows to fit a Gamma distribution function to the histogram. This can be seen from the red curve in Fig. \ref{synthetic_histo} and has been verified using a $\chi ^2$-test. In contrast to the measurements, the histograms are smoother. Especially, Fig. \ref{synthetic_histo} (a) has less outliers than the equivalent histogram of the measurements. The smooth shape is also an indicator to confirm the randomness of the generated data. However, to give a more precise evaluation of the generated data, we analyze the parameters of the Gamma distribution function that is fitted to the histograms.

\begin{figure}[tb]
\centering{\includegraphics[width=88mm]{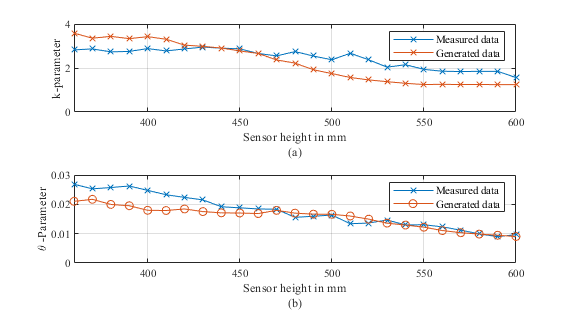}}
\caption{Shape $k$ (a) and scale $\theta$ (b) of the fitted Gamma distribution for different sensor heights, a beta angle of \SI{0}{\degree} on gravel for a range of \SI{0.5}{\meter} to \SI{0.75}{\meter} for measured data (blue) and generated data (red)}
\label{results}
\end{figure}

To validate the generated data clearly, we exemplarily choose the distance bin between \SI{0.5}{\meter} and \SI{0.75}{\meter} for a beta angle of \SI{0}{\degree}. The results for gravel are shown in Fig. \ref{results}. The parameters for the measured data are plotted in blue, whereas the parameters for the generated data are plotted in red. Both parameters have a high matching. Hence, the generated data is based on a very similar probability distribution. Consequently, the cGAN is able to model the ground reflections of ultrasonic signals precisely.

\begin{figure}[b]
\centering{\includegraphics[width=88mm]{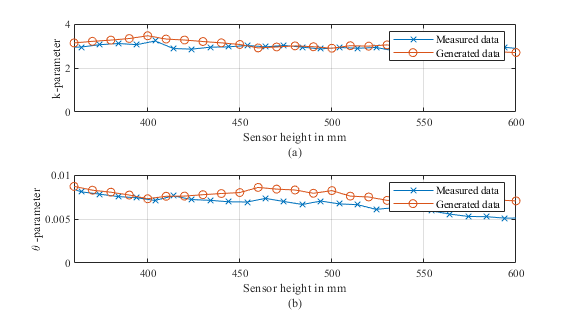}}
\caption{Shape $k$ (a) and scale $\theta$ (b) of the fitted Gamma distribution for different sensor heights, a beta angle of \SI{0}{\degree} on asphalt for a range of \SI{1.5}{\meter} to \SI{1.75}{\meter} for measured data (blue) and generated data (red)}
\label{results_asphalt}
\end{figure}

To further emphasize the high capability of cGANs to synthetically generate ultrasonic signals, the Gamma parameters of an asphalt scenario are evaluated. The data is generated with the same generator network. The change in the ground surface is just applied in the condition vector. The results for the distance bin at \SI{1.5}{\meter} to \SI{1.75}{\meter} are shown in Fig. \ref{results_asphalt}. The plotted curves of the parameters of the gamma distribution of the measured and generated signals again show a good matching. Hence, the network is able to handle also different ground types. Furthermore, the results show that the good matching of the parameters is not limited to a certain distance bin, but a generalization over all distance bins is achieved. This is important to not only create ground reflections at a particular delay. Consequently the proposed approach is able to generate realistic ultrasonic signals for complete environments without additional objects. 

%%%%%%%%%%%%%%%%%%%%%%%%%%%%%%%%%%%%%%%%%%%%%%%%%%%%%%%%%%%%%%%%%%%%%%%%%%%%%%%%

\section{Conclusion}
The use of GANs for synthetic image generation is well known. We adapted this approach for ultrasonic data for the first time. This novel approach enables a new level of realistic simulation of the environment in ultrasonic scenarios. To handle additional setup parameters, such as sensor height, installation angle or ground type, we used cGANs for data generation. We verified our approach by comparing the statistical properties of measurements and generated data. For all analyzed scenarios and parameters, the cGANs reach outstandingly realistic results. Hence, cGANs provide a extremely effective and fast method to provide a high amount of random ground reflections that are usable for ultrasonic signal simulation.

Due to the amazing performance of the proposed method, novel possibilities for the development of machine learning algorithms in automotive scenarios arise. Beside a speeding up of the algorithm training and evaluation, also online data generation is enabled. This allows an easy and time-effective refinement of methods. Furthermore there a new opportunities to calibrate sensor setups with respect to the environment online with a minimum time effort. To sum up, the proposed method is quantum leap towards an effective development of machine learning algorithms in ultrasonics and additionally opens up a new field of novel methods that require online data acquisition.


%%%%%%%%%%%%%%%%%%%%%%%%%%%%%%%%%%%%%%%%%%%%%%%%%%%%%%%%%%%%%%%%%%%%%%%%%%%%%%%%

%%%%%%%%%%%%%%%%%%%%%%%%%%%%%%%%%%%%%%%%%%%%%%%%%%%%%%%%%%%%%%%%%%%%%%%%%%%%%%%%

%%%%%%%%%%%%%%%%%%%%%%%%%%%%%%%%%%%%%%%%%%%%%%%%%%%%%%%%%%%%%%%%%%%%%%%%%%%%%%%%

\section*{ACKNOWLEDGMENT}

For his constructive comments and input, we thank Martin Simon and Johannes Petzold. Further we are grateful to Paul Rostocki for providing the ultrasonic sensors and Dr. Heinrich Gotzig for the technical supervising.

\bibliographystyle{IEEEtran}
\bibliography{IEEEabrv,biblio_traps_dynamics}

\end{document}